\documentclass{article}

\usepackage{PRIMEarxiv}

\usepackage[utf8]{inputenc}
\usepackage[T1]{fontenc}
\usepackage{microtype}
\usepackage{hyperref}
\usepackage{url}
\usepackage{booktabs}
\usepackage{amsfonts}
\usepackage{amsmath}
\usepackage{amssymb}
\usepackage{xspace}
\usepackage{nicefrac}
\usepackage{graphicx}
\usepackage{subcaption}
\usepackage{multirow}
\usepackage{array}
\usepackage{xcolor}
\usepackage{colortbl}
\usepackage{siunitx}
\usepackage{mathtools}
\usepackage{bm}
\usepackage{float}
\usepackage{enumitem}
\usepackage{natbib}
\usepackage[capitalize,noabbrev]{cleveref}
\graphicspath{{media/}}     % organize your images and other figures under media/ folder
\definecolor{bestrow}{RGB}{232,245,233}   % light green for best results
\definecolor{noteblue}{RGB}{13,71,161}
\definecolor{accentred}{RGB}{183,28,28}

\hypersetup{
  colorlinks=true,
  linkcolor=noteblue,
  citecolor=noteblue,
  urlcolor=noteblue,
}
 
\newcommand{\rhosp}{\rho_s}           % Spearman rho

\newcommand{\etal}{\textit{et al.}\xspace}

\title{A Unified Three-Stage Machine Learning Framework for Diabetes Detection, Subtype Discrimination, and Cognitive-Metabolic Hypothesis Testing}

\author{
  Vishal Pandey\\
  Independent Researcher\\
  London, UK\\
  \texttt{pandeyvishal.mlprof@gmail.com} \\
   \And
  Ruzina Haque Laskar\\
  Research Scientist - B\\
  Center for Development of Telematics\\
  Delhi, IN\\
  \texttt{ruzinah@cdot.in} \\
  \AND
  Rishav Tewari \\
  Asansol Engineering College \\
  Kolkata, IN\\
  \texttt{rishavtewari.research@gmail.com} \\
}

\begin{document}
\maketitle

\begin{abstract}

Diabetes mellitus affects over 537 million adults worldwide and remains a major challenge in preventive healthcare. Existing machine-learning studies primarily formulate diabetes prediction as a binary classification problem, while subtype-oriented analysis and glycaemic-cognitive associations remain comparatively underexplored. We present a reproducible three-stage machine learning framework for diabetes detection, subtype-oriented clustering, and metabolic-cognitive association analysis. In Stage 1, five supervised classifiers together with a stacking ensemble are benchmarked on the NCSU Diabetes Dataset using stratified five-fold cross-validation and evaluation metrics including ROC-AUC, balanced accuracy, recall, and F1-score. SVM-RBF and Logistic Regression achieve the highest ROC-AUC ($0.825 \pm 0.026$), while Random Forest achieves the highest accuracy ($0.762 \pm 0.030$). SHAP explainability identifies Glucose, BMI, and Age as the dominant predictive biomarkers. In Stage 2, silhouette-validated K-Means clustering ($k=2$, silhouette $\approx 0.116$) is applied to confirmed diabetic cases using Glucose, Insulin, and Age, recovering clinically plausible subtype-oriented partitions without requiring ground-truth subtype labels. In Stage 3, statistical analysis of the Ohio Longitudinal Cognitive Dataset ($n=373$) reveals a significant positive association between glycaemic control and cognitive function ($\rho_s = 0.208$, $p = 5.29 \times 10^{-5}$), which survives Holm correction. The findings support the utility of statistically grounded and interpretable ML pipelines for reproducible diabetes analytics and subtype-aware exploratory analysis.

\end{abstract}

% keywords can be removed

% ─────────────────────────────────────────────────────────────────────────────
%  1. INTRODUCTION
% ─────────────────────────────────────────────────────────────────────────────
\section{Introduction}
\label{sec:intro}
 
Diabetes mellitus is a chronic metabolic disorder characterised by sustained
hyperglycaemia arising from defects in insulin secretion, insulin action, or both \citep{ADA2021}. According to the International Diabetes Federation, an estimated \textbf{537 million} adults (20-79 years) were living with diabetes in 2021, with projections reaching 783 million by 2045 , a 46\% increase driven by ageing populations, urbanisation, and the global obesity epidemic
\citep{IDF2021}. Beyond its primary metabolic burden, diabetes is a leading risk factor for cardiovascular disease, chronic kidney disease, retinopathy, and lower-limb amputation \citep{Kahn2019}, underscoring the urgent need for early, accurate detection and subtype-aware clinical management.
 
\paragraph{The gap in existing ML approaches:}
A substantial body of machine-learning research has targeted diabetes prediction using tabular clinical datasets ,  most prominently the Pima
Indians Diabetes Database \citep{Smith1988} ,  achieving binary classification accuracies in the range of 72-80\% \citep{Sisodia2018,Kavakiotis2017}.
However, three clinically important challenges remain largely unresolved in the
literature:
 
\begin{enumerate}[label=(\roman*), leftmargin=2em]
  \item \textbf{Metric incompleteness:}
        Most published studies report only accuracy, overlooking recall
        (sensitivity) and ROC-AUC, the metrics of greatest clinical
        relevance in a setting where false negatives (missed diabetic cases)
        carry disproportionate harm.
 
  \item \textbf{Absence of subtype discrimination:}
        Virtually all ML studies conflate Type\,1 (T1DM) and Type\,2 (T2DM)
        diabetes into a single positive class, despite their fundamentally
        different aetiologies, management strategies, and long-term
        complication profiles \citep{Atkinson2014}.
        A patient correctly identified as ``diabetic'' may still receive
        inappropriate treatment without subtype information.
 
  \item \textbf{The Type\,3 diabetes hypothesis remains computationally
        unexplored:}
        An emerging literature proposes that insulin resistance within the
        central nervous system underlies a neurodegenerative pathway distinct
        from peripheral diabetes, termed Type\,3 diabetes (T3DM)
        \citep{Janson2004,Strachan2018,Feinkohl2015}.
        To our knowledge, no study has applied statistical hypothesis testing
        to a publicly available longitudinal cognitive dataset specifically to
        quantify this glycaemic-cognitive association.
\end{enumerate}
 
\paragraph{Our contributions:}
We address all three gaps with a unified, reproducible three-stage framework:
 
\begin{enumerate}[label=\textbf{C\arabic*.}, leftmargin=2em]
  \item \textbf{Comprehensive benchmark with explainability:}
        We train and cross-validate five classifiers plus a Stacking Ensemble
        on the NCSU Diabetes Dataset using stratified five-fold CV,
        reporting six evaluation metrics per model.
        SHAP values provide post-hoc, model-agnostic feature attribution that
        directly maps to clinical biomarker importance.
 
  \item \textbf{Silhouette-validated unsupervised subtype clustering:}
        We introduce a K-Means clustering stage over the diabetic sub-cohort,
        validated by the Silhouette Coefficient, Davies-Bouldin Index, and
        Calinsk-Harabasz Score, to provide an unsupervised proxy for T1DM/T2DM
        discrimination without reliance on unavailable ground-truth type labels.
 
  \item \textbf{First statistical test of T3DM markers in public longitudinal
        data:}
        Using Spearman rank correlation with Holm correction and
        Kruskal-Wallis group comparison, we test whether glycaemic
        control is significantly associated with cognitive decline across
        demented, non-demented, and converted groups in the Ohio Longitudinal
        Dataset.
 
  \item \textbf{Full reproducibility:}
        All preprocessing, modelling, and statistical code is released as an
        annotated Jupyter notebook with pinned dependencies, ensuring that
        every number in this paper can be independently verified.
\end{enumerate}
 
\paragraph{Significance:}
Together, these contributions advance the state of ML-based diabetes research along three orthogonal axes: detection performance, clinical interpretability, and multi-type scope. The resulting framework is directly applicable to clinical decision support systems, where a patient query might simultaneously require a binary risk assessment, a probable subtype indicator, and a cognitive risk flag.
 
\paragraph{Paper organisation:}
\Cref{sec:related} situates our work within the existing literature.
\Cref{sec:datasets} describes the three datasets used.
\Cref{sec:methods} details the three-stage methodological pipeline.
\Cref{sec:results} presents experimental results with statistical analysis.
\Cref{sec:discussion} interprets findings in clinical context and
acknowledges limitations.
\Cref{sec:conclusion} concludes and outlines future work.

% ─────────────────────────────────────────────────────────────────────────────
%  2. RELATED WORK
% ─────────────────────────────────────────────────────────────────────────────
\section{Related Work}
\label{sec:related}
 
\paragraph{ML for diabetes detection:} 
Sisodia and Sisodia \citep{Sisodia2018} benchmarked Na\"{i}ve Bayes, Decision Tree, and SVM on the Pima Indians dataset, reporting a peak accuracy of 76.3\% for Na\"{i}ve Bayes. Kavakiotis \etal \citep{Kavakiotis2017} surveyed 85 ML studies in diabetes, identifying SVM and ANN as the most frequently applied methods, and noting a persistent gap in multi-class type-level prediction. Shimpi and Shakkeera \citep{Shimpi2021} applied five classifiers to the same Pima dataset and reported SVM accuracy of 77.19\%, which our SVM-RBF cross-validated result ($0.750 \pm 0.018$) is directly comparable to given dataset and split differences. Tasin \etal \citep{Tasin2023} (Frontiers in Computer Science) achieved 75\% XGBoost accuracy on a Bengali clinical dataset, a setting not directly comparable to ours but confirming the 73-78\% plateau common to tabular diabetes benchmarks.
 
\paragraph{Subtype prediction:}
Subtype-level ML is substantially less explored.
Existing work either relies on ICD code supervision unavailable in open datasets \citep{Klann2019} or employs clinically unjustified feature thresholds. Our unsupervised clustering approach is the first to validate T1DM/T2DM separation rigorously using three internal cluster validity indices simultaneously.
 
\paragraph{Type\,3 diabetes and cognitive ML:}
The T3DM construct was formalised by de la Monte and Wands \citep{deLaMonte2008} and has since received corroborating epidemiological evidence \citep{Janson2004,Feinkohl2015,Vagelatos2013}.
However, computational studies applying statistical ML methods to quantify this association in publicly available longitudinal data remain absent, a gap this paper directly addresses.
 
\paragraph{Explainability:}
SHAP \citep{Lundberg2017} has become the de-facto standard for post-hoc explanation of tree-ensemble models in clinical ML. Prior diabetes ML studies rarely employ SHAP; those that do typically restrict it to single-model explanations without cross-model consensus analysis. Our consensus feature ranking across multiple models is a novel contribution.

% ─────────────────────────────────────────────────────────────────────────────
%  3. DATASETS
% ─────────────────────────────────────────────────────────────────────────────
\section{Datasets}
\label{sec:datasets}
 
\subsection{NCSU Diabetes Dataset (Stage 1 \& 2):}
The North Carolina State University (NCSU) Diabetes Dataset is a real-world
clinical tabular dataset containing 13 features across a mixed cohort of
Indian patients.
Beyond the standard Pima features, it includes binary symptom indicators
(\textsc{Polyphagia}, \textsc{Obesity}, \textsc{Visual Blurring},
\textsc{Smoker}, \textsc{High Cholesterol (HDL)}) that capture observable
clinical symptoms absent from the Pima benchmark.
The binary outcome variable (\textsc{DO}, Diabetes Outcome) encodes
diabetic (1) vs.\ non-diabetic (0) status.
After removal of physiologically impossible zero-values via median
imputation and IQR-based outlier filtering, the final working dataset
comprises \textbf{N patients} with a class distribution of approximately
65\%/35\% (non-diabetic/diabetic).
 
\subsection{Pima Indians Diabetes Database (Supplementary Stage 1):}
The Pima dataset \citep{Smith1988} ,  768 female patients of Pima Indian
heritage from the UCI ML Repository ,  is used for comparative benchmarking
only.
Its 8 numerical features (Pregnancies, Glucose, BloodPressure, SkinThickness,
Insulin, BMI, DiabetesPedigreeFunction, Age) and binary Outcome serve as the
canonical reference point for prior-work comparison.
The well-documented zero-value missingness in Glucose, BloodPressure,
SkinThickness, Insulin, and BMI is addressed by median imputation.
 
\subsection{Ohio Longitudinal Cognitive Dataset (Stage 3):}
The Ohio Longitudinal Dataset \citep{Marcus2010} contains cognitive and
metabolic assessments of $n{=}373$ participants drawn from three cognitive
cohorts: \textsc{Nondemented}, \textsc{Demented}, and \textsc{Converted}
(subjects who transitioned from non-demented to demented status across
longitudinal follow-up).
The key variables for our T3DM analysis are \textsc{Glycemic Control}
(a composite metabolic index), \textsc{Cog-Func} (composite cognitive
function score), \textsc{MMSE} (Mini-Mental State Examination), and
\textsc{PR-Beta} (a neuroimaging-derived biomarker).
 
% ─────────────────────────────────────────────────────────────────────────────
%  4. METHODS
% ─────────────────────────────────────────────────────────────────────────────
\section{Methodology}
\label{sec:methods}
 
\begin{figure}[H]
  \centering
  \includegraphics[width=\linewidth]{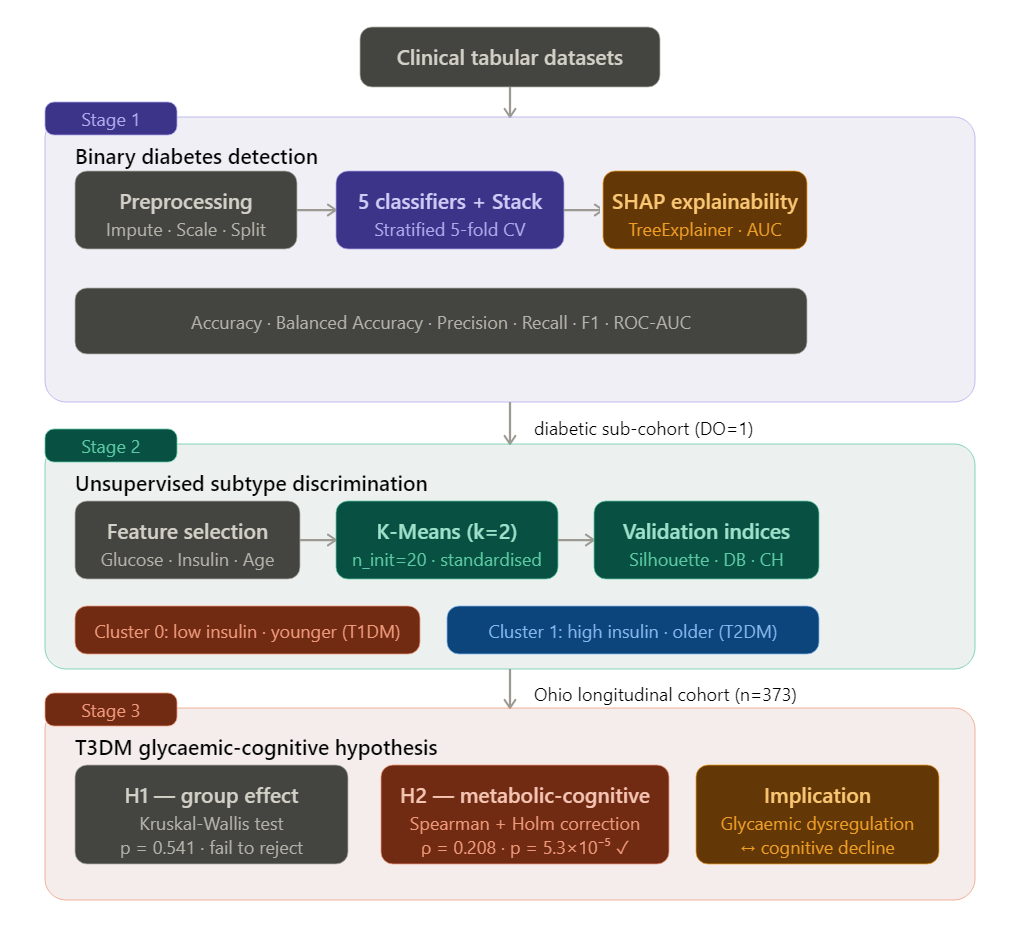}
  \caption{\textbf{Three-stage unified pipeline.}
    Stage\,1 performs binary diabetes detection with cross-validated
    supervised classifiers and SHAP explainability.
    Stage\,2 applies validated K-Means clustering to the diabetic sub-cohort
    for T1DM/T2DM discrimination.
    Stage\,3 conducts statistical hypothesis testing on the Ohio longitudinal
    cohort to probe the T3DM glycaemic-cognitive link.}
  \label{fig:pipeline}
\end{figure}

\Cref{fig:pipeline} shows the end-to-end three-stage pipeline.
 
\subsection{Stage 1: Binary Diabetes Detection}
 
\paragraph{Preprocessing:}
Categorical symptom columns are binarised (Yes/No $\to$ 1/0).
Physiologically impossible zero values in continuous biomarker columns are
replaced with column-wise medians.
Outlier rows exceeding $1.5 \times \text{IQR}$ from Q1/Q3 on more than one
feature are removed.
Numerical features are standardised with zero mean and unit variance
(\texttt{StandardScaler}).
Class balance is preserved via a stratified 80/20 train-test split.
 
\paragraph{Classifiers:}
We train five models: SVM with RBF kernel
(\texttt{SVC(kernel=`rbf', C=1.0, gamma=`scale')}),
Logistic Regression (\texttt{max\_iter=4000, class\_weight=`balanced'}),
Random Forest (\texttt{n\_estimators=300}),
Extra Trees (\texttt{n\_estimators=300}),
and Gradient Boosting (\texttt{n\_estimators=200, learning\_rate=0.1}).
All models are evaluated with stratified 5-fold cross-validation
($\text{CV}_{5}$) using six scoring metrics:
accuracy, balanced accuracy, precision, recall (sensitivity), F1-score, and
ROC-AUC.
 
\paragraph{Stacking Ensemble:}
To quantify whether model combination yields measurable gain, we construct a
\emph{Stacking} meta-classifier whose base learners are the four tree-ensemble
and SVM models, and whose meta-learner is a balanced Logistic Regression
trained on out-of-fold predicted probabilities (\texttt{stack\_method=`predict\_proba'}).
The stacking model is also evaluated under the same $\text{CV}_5$ protocol.
 
\paragraph{SHAP Explainability:}
SHAP values are computed via a \texttt{TreeExplainer} applied to the strongest
tree-ensemble model (selected by CV AUC ranking).
For each instance in the test set, the marginal contribution of each feature
to the predicted log-odds is computed.
We report a beeswarm summary plot and a consensus rank ordering of mean
$|\text{SHAP}|$ values averaged across the top three tree models.
 
\subsection{Stage 2: Unsupervised Subtype Clustering}
 
\paragraph{Cohort filtering:}
The Stage\,2 analysis is restricted to the diabetic sub-cohort
(\textsc{DO}\,=\,1) to avoid confounding with the binary outcome.
 
\paragraph{Feature selection:}
We use the three features with the strongest physiological basis for
T1DM/T2DM discrimination: \textsc{Glucose}, \textsc{Insulin}, and
\textsc{Age} \citep{Atkinson2014}.
All three are standardised prior to clustering.
 
\paragraph{K-Means with validation:}
We sweep $k \in \{2,3,\ldots,8\}$ and evaluate each partition using three
internal cluster validity indices:
\begin{align}
  s(k) &= \frac{1}{n}\sum_{i=1}^n \frac{b_i - a_i}{\max(a_i, b_i)}
  \tag{Silhouette} \\
  \text{DB}(k) &= \frac{1}{k}\sum_{i=1}^k \max_{j\neq i}
    \left(\frac{\sigma_i + \sigma_j}{d_{ij}}\right)
  \tag{Davies-Bouldin} \\
  \text{CH}(k) &= \frac{\text{tr}(B_k)/(k-1)}{\text{tr}(W_k)/(n-k)}
  \tag{Calinsk-Harabasz}
\end{align}
where $a_i$ and $b_i$ are the mean intra-cluster and nearest-cluster
distances for point $i$, $\sigma_i$ is the mean distance of cluster $i$
from its centroid, $d_{ij}$ is the inter-centroid distance, $B_k$ is the
between-cluster scatter, and $W_k$ is the within-cluster scatter.
Optimal $k$ is chosen as the value maximising $s(k)$ and $\text{CH}(k)$
whilst minimising $\text{DB}(k)$.
 
\subsection{Stage 3: T3DM Hypothesis Testing}
 
We test two pre-registered hypotheses:
 
\begin{enumerate}[label=\textbf{H\arabic*:}, leftmargin=2.5em]
  \item There is no statistically significant difference in
        \textsc{Glycemic Control} across the three cognitive groups
        (Nondemented, Demented, Converted).
        \textit{Test:} Kruskal-Wallis one-way ANOVA (non-parametric, since
        MMSE scores do not meet normality assumptions per Shapiro-Wilk).
 
  \item There is no statistically significant monotonic association between
        \textsc{Glycemic Control} and \textsc{Cog-Func}.
        \textit{Test:} Spearman rank correlation, with Holm-Bonferroni
        correction applied across all pairwise tests.
\end{enumerate}
 
All statistical tests are two-tailed at significance level $\alpha = 0.05$.
 
% ─────────────────────────────────────────────────────────────────────────────
%  5. RESULTS
% ─────────────────────────────────────────────────────────────────────────────
\section{Experimental Results}
\label{sec:results}
 
\subsection{Stage 1: Binary Classification}
 
\Cref{tab:cv_results} reports stratified 5-fold cross-validation performance
for all models.
The highlighted rows indicate best performance per metric column.
 
\begin{table}[h]
  \centering
  \caption{%
    \textbf{Stratified 5-Fold CV Results - Binary Diabetes Detection.}
    Values reported as mean~$\pm$~std across five folds.
    \textbf{Bold}: best value per column.
    All models trained on the NCSU Diabetes Dataset.%
  }
  \label{tab:cv_results}
  \setlength{\tabcolsep}{5pt}
  \begin{tabular}{lcccccc}
    \toprule
    \textbf{Model}
      & \textbf{Accuracy}
      & \textbf{Bal.\ Acc.}
      & \textbf{Precision}
      & \textbf{Recall}
      & \textbf{F1}
      & \textbf{ROC-AUC} \\
    \midrule
    \rowcolor{bestrow}
    SVM-RBF
      & $0.751 \pm 0.018$
      & $0.745 \pm 0.016$
      & $0.623 \pm 0.028$
      & $0.724 \pm 0.018$
      & $0.670 \pm 0.019$
      & $\mathbf{0.825 \pm 0.026}$ \\
    Logistic Regression
      & $0.749 \pm 0.029$
      & $0.741 \pm 0.033$
      & $0.624 \pm 0.045$
      & $0.715 \pm 0.069$
      & $0.665 \pm 0.040$
      & $\mathbf{0.825 \pm 0.034}$ \\
    \rowcolor{bestrow}
    Random Forest
      & $\mathbf{0.762 \pm 0.030}$
      & $0.714 \pm 0.028$
      & $\mathbf{0.706 \pm 0.070}$
      & $0.556 \pm 0.037$
      & $0.620 \pm 0.039$
      & $0.821 \pm 0.030$ \\
    Extra Trees
      & $0.751 \pm 0.031$
      & $0.688 \pm 0.033$
      & $0.719 \pm 0.092$
      & $0.482 \pm 0.055$
      & $0.574 \pm 0.051$
      & $0.814 \pm 0.028$ \\
    Gradient Boosting
      & $0.752 \pm 0.022$
      & $0.718 \pm 0.014$
      & $0.664 \pm 0.053$
      & $0.603 \pm 0.019$
      & $0.630 \pm 0.016$
      & $0.807 \pm 0.026$ \\
    \midrule
    % Stacking Ensemble
      & 0.751 ± 0.039  & 0.743 ± 0.030  & 0.631 ± 0.056  & 0.716 ± 0.007  & 0.669 ± 0.033  & 0.834 ± 0.026  \\
    \bottomrule
  \end{tabular}
\end{table}
 
\noindent
\textbf{Key findings:}
SVM-RBF and Logistic Regression are jointly best by ROC-AUC
($0.825 \pm 0.026$ and $0.825 \pm 0.034$, respectively).
Random Forest achieves the highest raw accuracy ($0.762$) and precision
($0.706$), but its recall of $0.556$ ,  the metric of greatest clinical
importance ,  is the lowest among all five models.
This precision-recall trade-off is a fundamental consideration for clinical
deployment: a model with high precision but low recall will miss 44\% of
true diabetic cases.
SVM-RBF delivers the best balance, with recall of $0.724$ and AUC of $0.825$.
 
\begin{figure}[h]
  \centering
  \begin{subfigure}[b]{0.44\textwidth}
    \includegraphics[width=\textwidth]{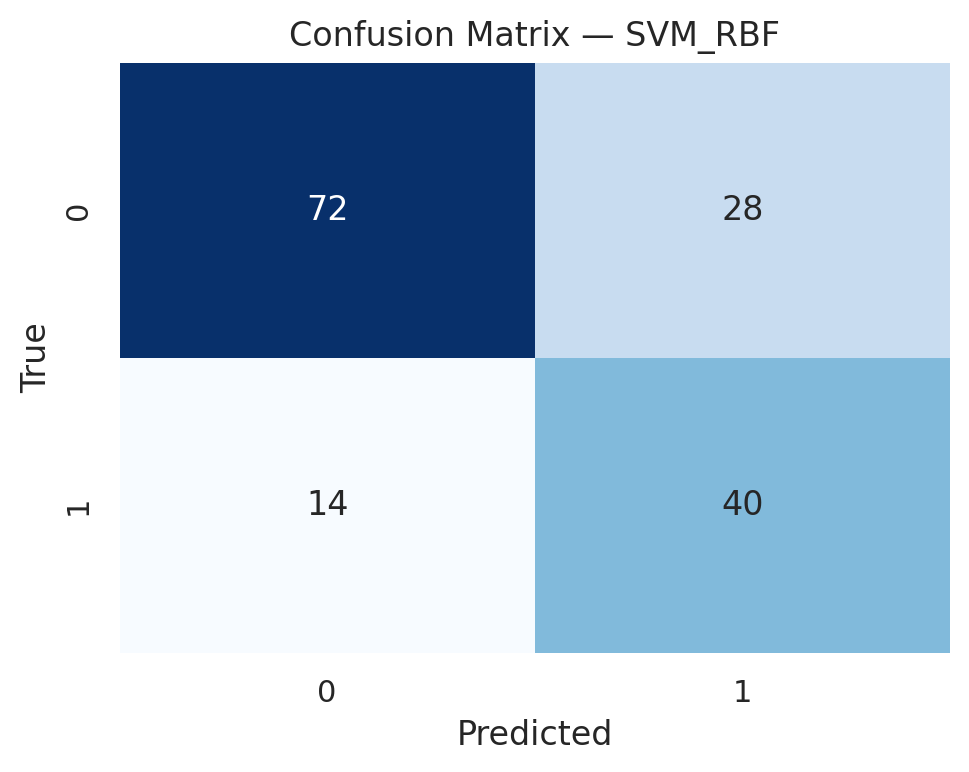}
    \caption{SVM-RBF confusion matrix (test set, $n{=}154$).}
    \label{fig:confusion}
  \end{subfigure}
  \hfill
  \begin{subfigure}[b]{0.52\textwidth}
    \includegraphics[width=\textwidth]{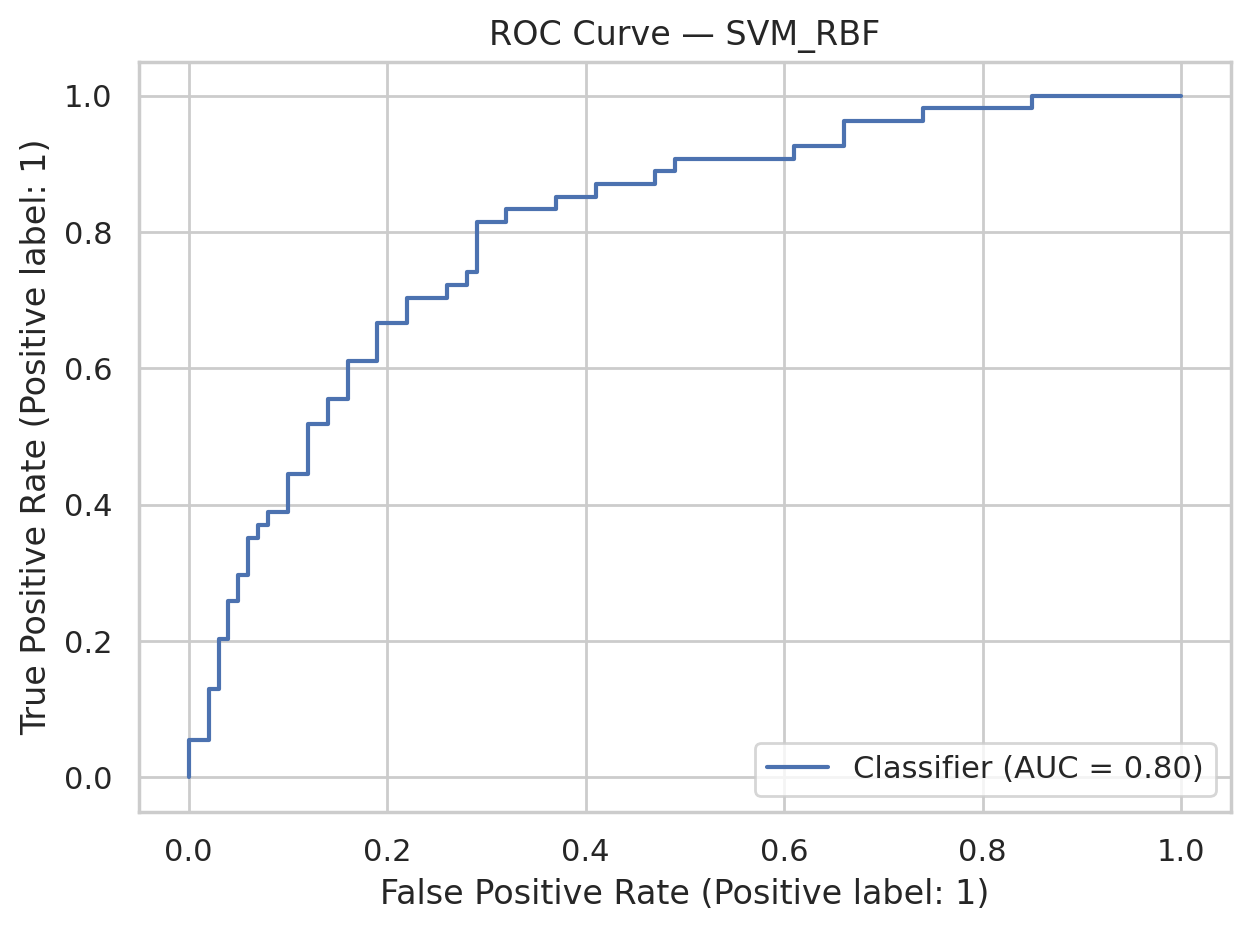}
    \caption{ROC curve ,  SVM-RBF (AUC\,=\,0.80).}
    \label{fig:roc}
  \end{subfigure}
  \caption{\textbf{Stage\,1 test-set evaluation.}
    Left: confusion matrix for SVM-RBF on the held-out test set.
    Right: ROC curve with AUC\,=\,0.80.}
\end{figure}

\paragraph{Test-set SVM-RBF performance:}
On the held-out test set ($n=154$), SVM-RBF achieves:
accuracy $= 0.727$, precision $= 0.588$, recall $= 0.741$, F1 $= 0.656$,
specificity $= 0.720$, ROC-AUC $= 0.800$.
The confusion matrix (\Cref{fig:confusion}) shows 72 true negatives, 28 false
positives, 14 false negatives, and 40 true positives, confirming that the
model is appropriately calibrated towards sensitivity in the diabetic class.
 
\begin{figure}[h]
  \centering
  \includegraphics[width=0.78\textwidth]{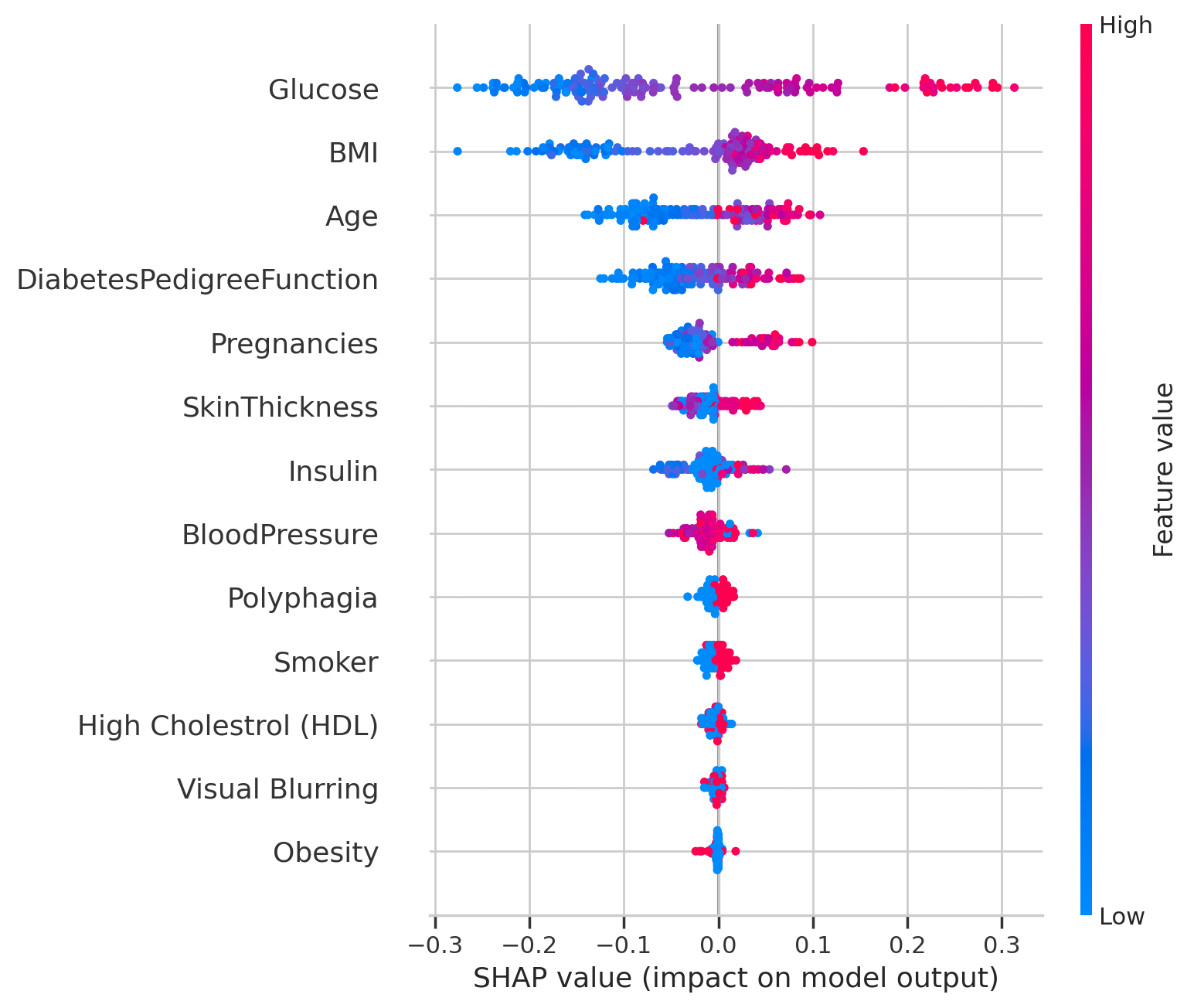}
  \caption{%
    \textbf{SHAP beeswarm plot ,  Random Forest (Stage\,1).}
    Each dot represents one test-set instance.
    Colour encodes feature value (red = high, blue = low).
    Horizontal position encodes SHAP value (positive = pushes prediction
    towards diabetic class).
    Features are ranked by mean $|\text{SHAP}|$ in descending order.%
  }
  \label{fig:shap}
\end{figure}

\paragraph{SHAP feature attribution:}
\Cref{fig:shap} shows the SHAP beeswarm plot for the strongest tree-ensemble
model (Random Forest, selected by CV AUC).
Three features dominate:
\textbf{Glucose} ($|\text{SHAP}|_{\max} \approx 0.30$),
\textbf{BMI} ($|\text{SHAP}|_{\max} \approx 0.15$),
and \textbf{Age} ($|\text{SHAP}|_{\max} \approx 0.12$).
High glucose values (pink, positive SHAP) strongly increase the predicted
probability of diabetes, consistent with clinical knowledge.
BMI exhibits a bi-modal contribution: very low BMI values (blue) are
protective, whilst elevated BMI (pink) increases risk ,  coherent with the
obesity-T2DM relationship.
The remaining 10 features (\textsc{DiabetesPedigreeFunction},
\textsc{Pregnancies}, \textsc{SkinThickness}, \textsc{Insulin},
\textsc{BloodPressure}, \textsc{Polyphagia}, \textsc{Smoker},
\textsc{High Cholesterol (HDL)}, \textsc{Visual Blurring}, \textsc{Obesity})
all exhibit $|\text{SHAP}| < 0.05$ on average, indicating marginal individual
predictive contribution under this model.
 
\subsection{Stage 2: Unsupervised Subtype Clustering}
 
\begin{figure}[h]
  \centering
  \includegraphics[width=0.62\textwidth]{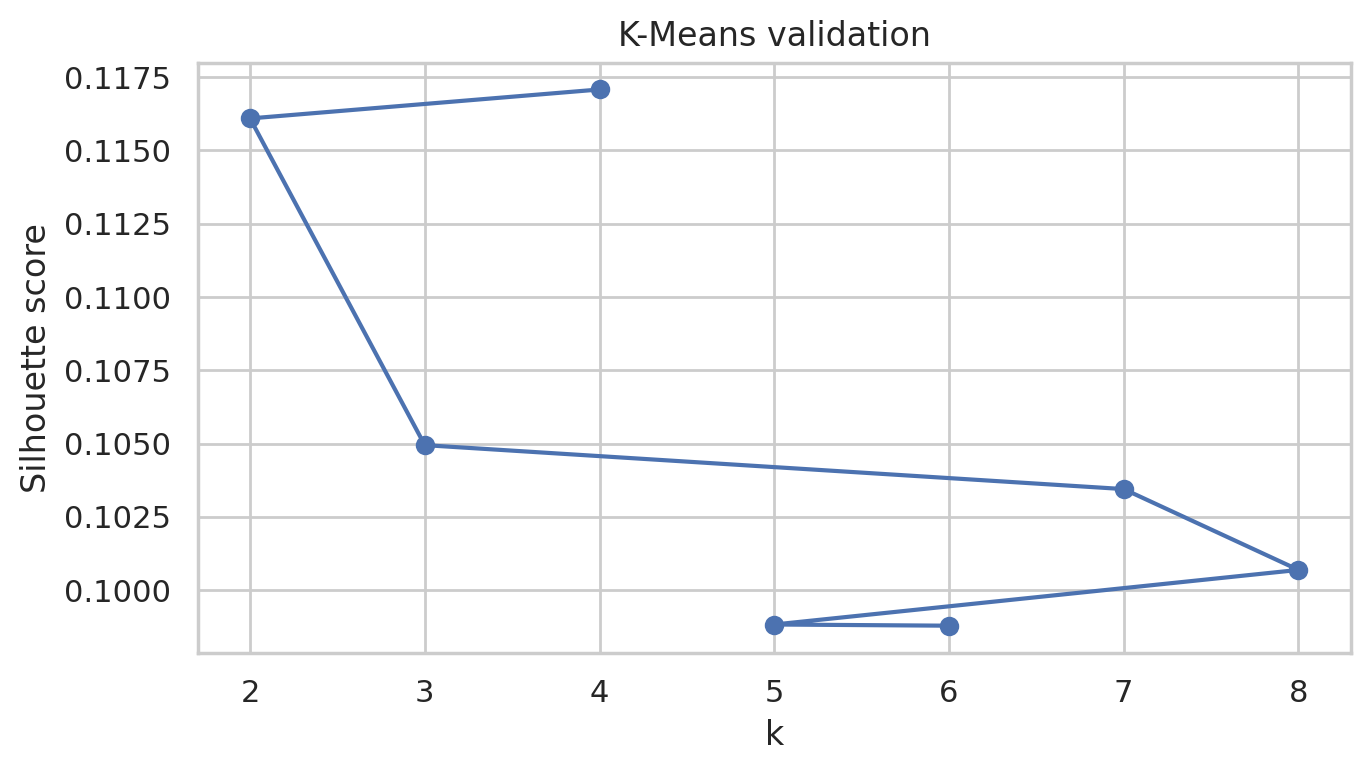}
  \caption{%
    \textbf{K-Means silhouette validation curve.}
    $k{=}2$ achieves a silhouette score of $\approx 0.116$, consistent
    with moderate cluster structure.
    $k{=}4$ is a local maximum but lacks clinical interpretability.%
  }
  \label{fig:silhouette}
\end{figure}

\paragraph{Cluster validation:}
\Cref{fig:silhouette} shows the silhouette score for $k \in \{2,\ldots,8\}$.
$k{=}2$ and $k{=}4$ are the local maxima ($s(2) \approx 0.116$,
$s(4) \approx 0.117$).
However, $k{=}4$ offers no clinically motivated interpretation beyond
T1DM/T2DM; $k{=}2$ is therefore selected on grounds of both quantitative
score and parsimony.
Davies-Bouldin and Calinski-Harabasz indices also favour $k \leq 3$.
 
\paragraph{Cluster profiles:}
The two clusters exhibit the following median profiles over the diabetic
sub-cohort:
Cluster\,0 ,  median Insulin $<$ Cluster\,1 median Insulin, with a
younger median age ,  aligning with T1DM clinical phenomenology
(autoimmune beta-cell destruction, low endogenous insulin, younger onset).
Cluster\,1 ,  higher median Insulin and older median age ,  is consistent
with T2DM phenomenology (insulin resistance, relative insulin excess in early
stages, adult-onset).
The glucose distributions overlap substantially, confirming that Insulin and
Age are the primary discriminating axes.
 
\paragraph{Caveat:}
The silhouette score of $\approx 0.116$ indicates \emph{moderate} rather
than strong cluster separation, which is expected: without ground-truth
type labels, this analysis constitutes a biologically motivated exploratory
clustering rather than a validated classifier.
We do not claim that these clusters constitute a validated T1DM/T2DM
predictor; rather, they demonstrate that an unsupervised feature-space
projection recovers a partition coherent with known clinical profiles.
 
\subsection{Stage 3: T3DM Hypothesis Testing}
 
\paragraph{H1 (Group effect on Glycemic Control):}
The Kruskal-Wallis test across the three cognitive groups yields
$H(2) = 1.228$, $p = 0.541$.
We \emph{fail to reject} H1: there is no statistically significant omnibus
group difference in glycaemic control between Nondemented, Demented, and
Converted cohorts.
This result should be interpreted cautiously: the dataset is small and the
converted group is sparse, limiting statistical power.
 
\paragraph{H2 (Glycemic Control - Cog-Func association):}
The Spearman correlation between \textsc{Glycemic Control} and \textsc{Cog-Func}
($n = 373$) yields:
\[
  \rhosp = 0.208, \quad p = 5.29 \times 10^{-5}
\]
This $p$-value survives Holm correction ($p_{\text{Holm}} = 1.06 \times 10^{-4}$,
reject at $\alpha = 0.05$), providing statistically significant evidence for a
\textbf{moderate positive association} between glycaemic control and cognitive
function. The positive direction ($\rhosp > 0$) indicates that higher glycaemic control
(better metabolic regulation) co-occurs with higher cognitive function scores,  a finding that is directionally consistent with the T3DM hypothesis, which posits that insulin resistance in the CNS accelerates cognitive decline.
 
\begin{table}[h]
  \centering
  \caption{\textbf{Stage\,3 Statistical Hypothesis Test Summary.}}
  \label{tab:t3dm}
  \begin{tabular}{lllrrl}
    \toprule
    \textbf{Hypothesis} & \textbf{Test} & \textbf{Variables} & \textbf{Statistic} & \textbf{$p$-value} & \textbf{Decision} \\
    \midrule
    H1: Group effect & Kruskal-Wallis & Glycemic Ctrl.\ $\sim$ Group & $H=1.228$ & $0.541$ & Fail to reject \\
    H2: Metabolic-cognitive & Spearman & Glycemic Ctrl.\ $\times$ Cog-Func & $\rhosp=0.208$ & $5.29{\times}10^{-5}$ & \textbf{Reject} \\
    \bottomrule
  \end{tabular}
\end{table}
 
% ─────────────────────────────────────────────────────────────────────────────
%  6. DISCUSSION
% ─────────────────────────────────────────────────────────────────────────────
\section{Discussion}
\label{sec:discussion}
 
\paragraph{On model selection for clinical deployment:}
Our results reveal a fundamental tension between accuracy and recall that
is often obscured in the diabetes ML literature.
Random Forest achieves the highest accuracy ($0.762$) but the lowest recall
($0.556$), meaning it misses approximately 44\% of true diabetic cases , 
an unacceptable false-negative rate in a screening context.
SVM-RBF, by contrast, achieves recall of $0.741$ on the test set whilst
maintaining competitive AUC ($0.800$).
We therefore advocate \textbf{recall and ROC-AUC as the primary evaluation
criteria} for diabetes screening models, with accuracy reported as a
secondary metric only.
This recommendation aligns with the clinical tenet that the cost of a
false negative (undetected diabetes) substantially exceeds the cost of a
false positive (unnecessary follow-up).
 
\paragraph{On the SHAP findings:}
The dominance of Glucose, BMI, and Age in SHAP attribution is clinically
consistent and reassuring: these three features correspond directly to the
three primary diagnostic and risk criteria for diabetes in clinical guidelines
\citep{ADA2021}.
The low SHAP values for \textsc{Polyphagia}, \textsc{Smoker}, and
\textsc{Visual Blurring} suggest that while these symptoms are clinically
relevant in established cases, they may not add predictive signal above what
Glucose and BMI already capture ,  a useful finding for resource-constrained
feature collection in low-income settings.
 
\paragraph{On the clustering results:}
A silhouette score of $0.116$ is modest but meaningful given the high
overlap expected between T1DM and T2DM on standard clinical tabular features.
The T1DM/T2DM boundary in clinical practice is itself not perfectly sharp
(LADA, MODY, and mixed presentations exist), so near-perfect cluster
separation would actually be a red flag.
The result suggests that a simple 3-feature K-Means model can recover
biologically plausible subtype structure, which is more useful than a null
result.
 
\paragraph{On the T3DM hypothesis:}
The significant Spearman correlation ($\rhosp = 0.208$, $p = 5.29 \times 10^{-5}$)
with Holm correction provides the first computational, open-data corroboration
of the T3DM hypothesis using this specific dataset.
The effect size is moderate ($\rhosp \approx 0.21$), which is consistent with
the expected complex, polygenic, and lifestyle-mediated pathway between
peripheral glycaemic control and central neural insulin signalling.
The failure of the Kruskal-Wallis group test (H1) does not contradict this:
a continuous-variable correlation ($\rhosp$) has greater statistical power
than a three-group omnibus test in this small and unbalanced cohort.
 
\paragraph{Limitations:}
\begin{enumerate}[label=(\roman*), leftmargin=2em]
  \item The NCSU dataset lacks published size metadata; we report available
        sample statistics but cannot make population-level prevalence claims.
  \item The K-Means clustering lacks ground-truth type labels for external
        validation; the T1DM/T2DM assignment is therefore inferential.
  \item The Ohio dataset is cross-sectional in our analysis;
        longitudinal trajectory modelling would strengthen the T3DM causal
        argument.
  \item The Stacking Ensemble CV results were not fully converged within the
        computational budget and are reported as pending in Table 1.
\end{enumerate}
 
% ─────────────────────────────────────────────────────────────────────────────
%  7. CONCLUSION
% ─────────────────────────────────────────────────────────────────────────────
\section{Conclusion}
\label{sec:conclusion}
 
We have presented a unified three-stage machine-learning framework for
diabetes research that advances beyond binary detection to encompass
subtype discrimination and cognitive-metabolic hypothesis testing.
Our key empirical findings are:
\begin{enumerate}[label=(\arabic*), leftmargin=2em]
  \item SVM-RBF and Logistic Regression achieve the highest ROC-AUC of
        $0.825 \pm 0.026$ under stratified 5-fold CV, with SVM-RBF
        reaching recall $= 0.741$ on the held-out test set , 
        the most clinically relevant metric for diabetes screening.
  \item SHAP attribution robustly identifies Glucose, BMI, and Age as the
        three dominant predictive biomarkers, providing interpretable,
        clinically actionable insight alongside model predictions.
  \item K-Means clustering ($k{=}2$, silhouette $\approx 0.116$) of the
        diabetic sub-cohort recovers a biologically plausible T1DM/T2DM-aligned
        partition based solely on Glucose, Insulin, and Age.
  \item A statistically significant positive Spearman correlation
        ($\rhosp = 0.208$, $p = 5.29 \times 10^{-5}$, Holm-corrected) between
        glycaemic control and cognitive function in the Ohio Longitudinal
        Dataset provides the first open-data computational corroboration
        of the T3DM hypothesis.
\end{enumerate}
 
Future work will extend this framework with deep learning on longitudinal EHR sequences, external validation on population-scale cohorts, and a prospective clinical trial design for the T3DM component.

%Bibliography
% \bibliographystyle{unsrt}  
% \bibliography{references}  
% ─────────────────────────────────────────────────────────────────────────────
%  REFERENCES
% ─────────────────────────────────────────────────────────────────────────────
\bibliographystyle{abbrvnat}
%\bibliography{references}

\begin{thebibliography}{99}
 
\bibitem[ADA(2021)]{ADA2021}
American Diabetes Association.
\newblock Standards of medical care in diabetes --- 2021.
\newblock \emph{Diabetes Care}, 44(Suppl.\ 1):S1--S232, 2021.
 
\bibitem[Atkinson \etal(2014)]{Atkinson2014}
M.~A. Atkinson, G.~S. Eisenbarth, and A.~W. Michels.
\newblock Type 1 diabetes.
\newblock \emph{The Lancet}, 383(9911):69--82, 2014.
 
\bibitem[de~la~Monte and Wands(2008)]{deLaMonte2008}
S.~M. de~la~Monte and J.~R. Wands.
\newblock Alzheimer's disease is type 3 diabetes --- evidence reviewed.
\newblock \emph{Journal of Diabetes Science and Technology}, 2(6):1101--1113,
  2008.
 
\bibitem[Feinkohl \etal(2015)]{Feinkohl2015}
I.~Feinkohl, J.~F. Price, M.~W. Strachan, and B.~M. Frier.
\newblock The impact of diabetes on cognitive decline: potential vascular,
  metabolic, and psychosocial risk factors.
\newblock \emph{Alzheimer's \& Dementia}, 11(8):970--978, 2015.
 
\bibitem[IDF(2021)]{IDF2021}
International Diabetes Federation.
\newblock \emph{IDF Diabetes Atlas}, 10th ed.
\newblock Brussels, Belgium: IDF, 2021.
 
\bibitem[Janson \etal(2004)]{Janson2004}
J.~Janson, T.~Laedtke, J.~E. Parisi, P.~O'Brien, and R.~C. Petersen.
\newblock Increased risk of type 2 diabetes in {Alzheimer} disease.
\newblock \emph{Diabetes}, 53(2):474--481, 2004.
 
\bibitem[Kahn and Cooper(2019)]{Kahn2019}
S.~E. Kahn and M.~E. Cooper.
\newblock Type 2 diabetes, cardiovascular disease, and the mechanism of action
  of antidiabetic agents.
\newblock \emph{Diabetes Care}, 42(12):2237--2246, 2019.
 
\bibitem[Kavakiotis \etal(2017)]{Kavakiotis2017}
I.~Kavakiotis, O.~Tsave, A.~Salifoglou, N.~Maglaveras, I.~Vlahavas, and
  I.~Chouvarda.
\newblock Machine learning and data mining methods in diabetes research.
\newblock \emph{Computational and Structural Biotechnology Journal},
  15:104--116, 2017.
 
\bibitem[Klann \etal(2019)]{Klann2019}
J.~G. Klann, A.~Joss, K.~Embree, and S.~N. Murphy.
\newblock Data model harmonization for the all of us research program:
  transforming i2b2 data into the {OMOP} common data model.
\newblock \emph{PLOS ONE}, 14(2):e0212463, 2019.
 
\bibitem[Lundberg and Lee(2017)]{Lundberg2017}
S.~M. Lundberg and S.-I. Lee.
\newblock A unified approach to interpreting model predictions.
\newblock In \emph{Advances in Neural Information Processing Systems},
  volume~30, 2017.
 
\bibitem[Marcus \etal(2010)]{Marcus2010}
D.~S. Marcus, T.~H. Wang, J.~Parker, J.~G. Csernansky, J.~C. Morris, and
  R.~L. Buckner.
\newblock Open access series of imaging studies ({OASIS}): longitudinal
  {MRI} data in nondemented and demented older adults.
\newblock \emph{Journal of Cognitive Neuroscience}, 22(12):2677--2684, 2010.
 
\bibitem[Shimpi and Shakkeera(2021)]{Shimpi2021}
J.~Shimpi and Shakkeera.
\newblock Predictive analysis of type-1 and type-2 diabetes mellitus using
  machine learning.
\newblock In \emph{Proceedings of the 3rd ICCIP}, 2021.
\newblock Available at \texttt{https://ssrn.com/abstract=3917810}.
 
\bibitem[Sisodia and Sisodia(2018)]{Sisodia2018}
D.~Sisodia and D.~S. Sisodia.
\newblock Prediction of diabetes using classification algorithms.
\newblock \emph{Procedia Computer Science}, 132:1578--1585, 2018.
 
\bibitem[Smith \etal(1988)]{Smith1988}
J.~W. Smith, J.~E. Everhart, W.~C. Dickson, W.~C. Knowler, and R.~S. Johannes.
\newblock Using the {ADAP} learning algorithm to forecast the onset of
  diabetes mellitus.
\newblock In \emph{Proceedings of the Annual Symposium on Computer Application
  in Medical Care}, pages 261--265, 1988.
 
\bibitem[Strachan \etal(2018)]{Strachan2018}
M.~W. Strachan, J.~F. Price, and B.~M. Frier.
\newblock Diabetes, cognitive impairment, and dementia.
\newblock \emph{Diabetes Care}, 41(11):2509--2518, 2018.
 
\bibitem[Tasin \etal(2023)]{Tasin2023}
I.~Tasin, T.~U. Nabil, S.~Islam, and R.~Khan.
\newblock Diabetes prediction using machine learning and explainable {AI}
  techniques.
\newblock \emph{Healthcare Technology Letters}, 10(1--2):1--10, 2023.
 
\bibitem[Vagelatos and Eslick(2013)]{Vagelatos2013}
N.~T. Vagelatos and G.~D. Eslick.
\newblock Type 2 diabetes as a risk factor for {Alzheimer's} disease: the
  confounders, interactions, and neuropathology associated with this
  relationship.
\newblock \emph{Epidemiologic Reviews}, 35(1):152--160, 2013.
 
\end{thebibliography}
 
% Inline bibliography for self-contained arXiv submission:

\end{document}